\documentclass[conference]{IEEEtran}

\makeatletter
\def\footnoterule{\relax%
  \kern-5pt
  \hbox to \columnwidth{\hfill\vrule width 0.8\columnwidth height 0.4pt\hfill}
  \kern4.6pt}
\makeatother
\usepackage{amsmath}
\usepackage{graphicx}
\usepackage{algorithm2e}
\usepackage{siunitx} 

\usepackage{multirow}

\usepackage{xcolor}
\definecolor{redcolor}{rgb}{1, 0, 0}

\IEEEoverridecommandlockouts
\begin{document}
%
\title{Unsupervised Competitive Hardware Learning Rule for Spintronic Clustering Architecture}

\author{\IEEEauthorblockN{Alvaro Velasquez$^1$, Christopher H. Bennett\thanks{DISTRIBUTION STATEMENT A.  Approved for public release: distribution is unlimited. Approval ID: 88ABW-2019-4744. (Date of determination: September 27, 2019). This paper describes objective technical results and analysis. Any subjective views or opinions that might be expressed in the paper do not necessarily represent the views of the U.S. Air Force, U.S. Department of Energy, or the United States Government. Sandia National Laboratories is a multimission laboratory managed and operated by National Technology \& Engineering Solutions of Sandia, LLC, a wholly owned subsidiary of Honeywell International Inc., for the U.S. Department of Energy's National Nuclear Security Administration under contract DE-NA0003525.}$^2$, Naimul Hassan$^3$, Wesley H. Brigner$^3$, Otitoaleke G. Akinola$^4$,\\
Jean Anne C. Incorvia$^4$, Matthew J. Marinella$^2$, Joseph S. Friedman$^3$}
\IEEEauthorblockA{$^1$Information Directorate, Air Force Research Laboratory, Rome, NY 13441\\
$^2$Sandia National Laboratories, Albuquerque, NM 87123\\
$^3$Department of Electrical and Computer Engineering, University of Texas at Dallas, Richardson, TX 75080\\
$^4$Department of Electrical and Computer Engineering, University of Texas at Austin, Austin, TX 78712\\
alvaro.velasquez.1@us.af.mil, \{cbennet, mmarine\}@sandia.gov,\\ \{naimul.hassan, wesley.brigner, joseph.friedman\}@utdallas.edu, otitoaleke@utexas.edu, incorvia@austin.utexas.edu}}

%
%
%
%

%

\setlength{\parskip}{0pt}
\setlength{\topsep}{0pt}
\setlength{\partopsep}{0pt}

\setlength{\abovedisplayskip}{0pt}
\setlength{\belowdisplayskip}{0pt}

\setlength{\floatsep}{0pt} 
\setlength{\textfloatsep}{0pt} 
\setlength{\intextsep}{0pt} 
\setlength{\dbltextfloatsep}{0pt} 
\setlength{\dblfloatsep}{0pt} 
\setlength{\abovecaptionskip}{0pt} 
\setlength{\belowcaptionskip}{0pt} 


\maketitle


\begin{abstract}
We propose a hardware learning rule for unsupervised clustering within a novel spintronic computing architecture. The proposed approach leverages the three-terminal structure of domain-wall magnetic tunnel junction devices to establish a feedback loop that serves to train such devices when they are used as synapses in a neuromorphic computing architecture.
\end{abstract}

%

\section{Introduction}
Neuromorphic computing promises exceptional capabilities for artificial intelligence through highly-efficient circuit structures mimicking the structure and functionality of the human brain. While neural networks composed of artificial synapses and neurons have been demonstrated with conventional CMOS devices, it is expected that computing efficiency can be improved by more closely emulating neurobiological hardware. Thus, the potential use of non-volatile spintronic analog devices that can be interconnected in a manner that enables on-chip unsupervised learning is particularly promising. In particular, spintronic domain-wall (DW) devices have been proposed as capable synapses \cite{DWSynapse} and neurons \cite{DWNeuron} and achieve a remarkable energy efficiency and reliability relative to competing non-volatile candidate devices such as resitive/filamentary memristive and phase-change memory (PCM) candidate devices for on-chip learning \cite{kaushik2019comparing}. Recently, DW synaptic devices have been co-integrated with transistor learning circuits to show basic learning \cite{yue2019brain}.

Domain wall-magnetic tunnel junction devices (DW-MTJs) are spintronic devices that provide non-volatile memory. The position of the DW along its track determines the tunneling resistance of a DW-MTJ and can be modulated by applying a current through the track. Three-terminal DW-MTJs have been leveraged as analog artificial synapses with long tunnel barriers and as binary artificial neurons by a digital version with shorter tunnel barrriers \cite{hassanJAP}. Furthermore, a CMOS-free monolothic multi-layer perceptron system can be achieved solely with three-terminal DW-MTJ synapses and four-terminal DW-MTJ neurons intrinsically capable of emulating the leaky-integrate-and-fire (LIF) neuron model. However, as this perceptron system does not utilize the third terminal of the DW-MTJ synapses, there is an opportunity to use this third terminal for efficient on-chip unsupervised learning. We propose such an approach to clustering using DW-MTJ crossbars that minimize the use of CMOS. 



\begin{figure}
    \centering
    \includegraphics[width=5cm]{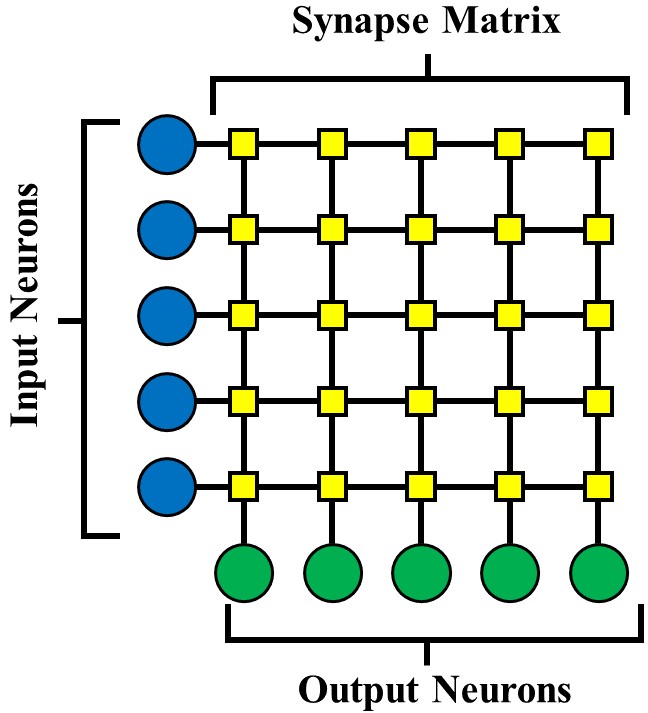} 
    \caption{A $5\times5$ neuromorphic crossbar array.}
    \label{fig:Crossbar_Array}
\end{figure}

\section{Neuromorphic Crossbar Array: Background}
A neuromorphic crossbar array is a hardware organization of circuit units that can perform brain-inspired functionalities, and has its origins in the use of crossbars for logic and memory. Like the crossbar array, the biological brain can be simplified and abstracted as a neural network consisting of neurons connected by synapses.

Neurons receive electrical or chemical stimuli through the dendrites, perform computations in the cell body (\textit{i.e.}, soma), and propagate the output through axons. Synapses are the junctions between neurons that modulate the signal propagation strength. The leaky integrate-and-fire (LIF) neuron is a popular artificial neuron model that mimics the biological neuron. An LIF neuron continuously integrates energy provided through the synapses connected to the input neuron.Throughout the integration process, the stored energy leaks if the energy received by the neuron is insufficient. When the stored energy reaches a particular threshold, the LIF neuron generates a spike that can be propagated to the next connected neuron or be utilized to perform inference. 

The synaptic connections of a neuromorphic crossbar can be implemented with variable resistors and many such crossbar architectures have been proposed using memristors \cite{memristorSynapse}, 1-transistor 1-resistor RRAM \cite{RRAM}, phase-change memories \cite{PCM}, and ferroelectric RAM \cite{ferroelectric}, among others. These variable resistors define the weighted connections between neurons. The weight tuning is performed by changing the conductance of the synapses and is referred to as `learning' when the tuning is conducive to the optimization of some objective function for which a cognitive task is defined (i.e. face recognition).

Fig. \ref{fig:Crossbar_Array} shows a $5\times5$ neuromorphic crossbar array where horizontal word lines are connected to the input neurons and vertical bit lines are connected to the output neurons. At the intersection of each word line and bit line, a synapse is placed that defines the connectivity between the input and the output neuron. In this fashion, a weight matrix is formed by the synapses, and the output neurons receive the product of vector-matrix multiplication between an input vector and the weight matrix. Cascading multiple crossbar array results in a multi-layer perceptron.

\begin{figure}
    \centering
    \includegraphics[width=\columnwidth]{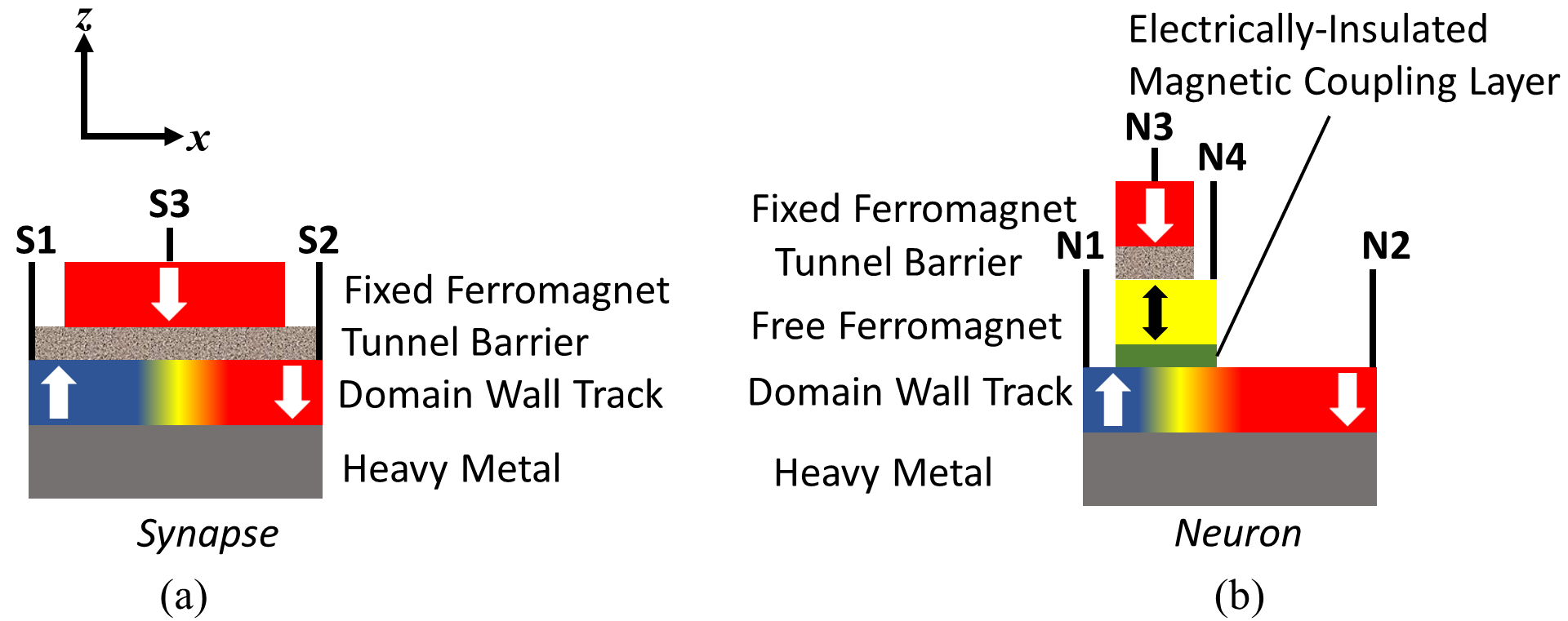} 
    \caption{(a) A 3T DW-MTJ synapse which has an analog conductance determined by the  position of the DW. (b) A 4T DW-MTJ neuron which provides binary conductance between the N3-N4 path, and can perform LIF functionalities.}
    \label{fig:Synapse_Neuron}
\end{figure}

\section{DW-MTJ Synapse and Neuron}
\label{sec:Synapse_Neuron}
Fig. \ref{fig:Synapse_Neuron}(a) shows a three-terminal (3T) DW-MTJ that acts as a synapse by providing an analog conductance element to the neural system \cite{dutta2017logic}. Between terminals S1 and S2, it has a free ferromagnetic domain wall track divided into +\textit{z} and -\textit{z} directed magnetic domains separated by a DW. An analog MTJ is formed between the S1 and S3 terminals by the DW track, tunnel barrier and, the  -\textit{z} directed fixed ferromagnet. The position of the DW can be controlled by applying the S2-S1 spin-transfer torque inducing write current. The MTJ conductance can be represented as a parallel-resistance combination of parallel and anti-parallel (with respect to the fixed ferromagnet) portions of the DW-MTJ. A positive (negative) S2-S1 current grows the anti-parallel (parallel) +\textit{z} (-\textit{z}) domain and decreases (increases) the conductance of the DW-MTJ. 

Fig. \ref{fig:Synapse_Neuron}(b) shows a four-terminal (4T) version of DW-MTJ which performs the leaky integrate-and-fire neuronal functionalities \cite{brigner4TNeuron2020}. Similar to the synapse, it has a DW track between terminals N1 and N2. The magnetic state of the DW track is coupled to a free ferromagnet of smaller dimension. The electrically-insulating magnetic coupling layer separates the N1-N2 write current path from the N3-N4 read path. The N3-N4 path has an MTJ of smaller dimension compared to the synapse MTJ. Spin-transfer torque inducing current through the N1-N2 path performs integration by moving the DW from right-to-left direction. Leaking causes the DW to move in left-to-right direction and can be performed by one of three approaches - dipolar coupling field \cite{hassanJAP}, anisotropy gradient along the DW track \cite{brignerJxCDC} and shape variation of the DW track \cite{brignerTED}. When the DW crosses beneath the dipolarly-coupled free ferromagnet during the integration process, the MTJ between the N3-N4 path switches from an anti-parallel low conductive state to a parallel high conductive state, thus performing the firing functionality.

\begin{figure}
    \centering
    \includegraphics[width=8cm]{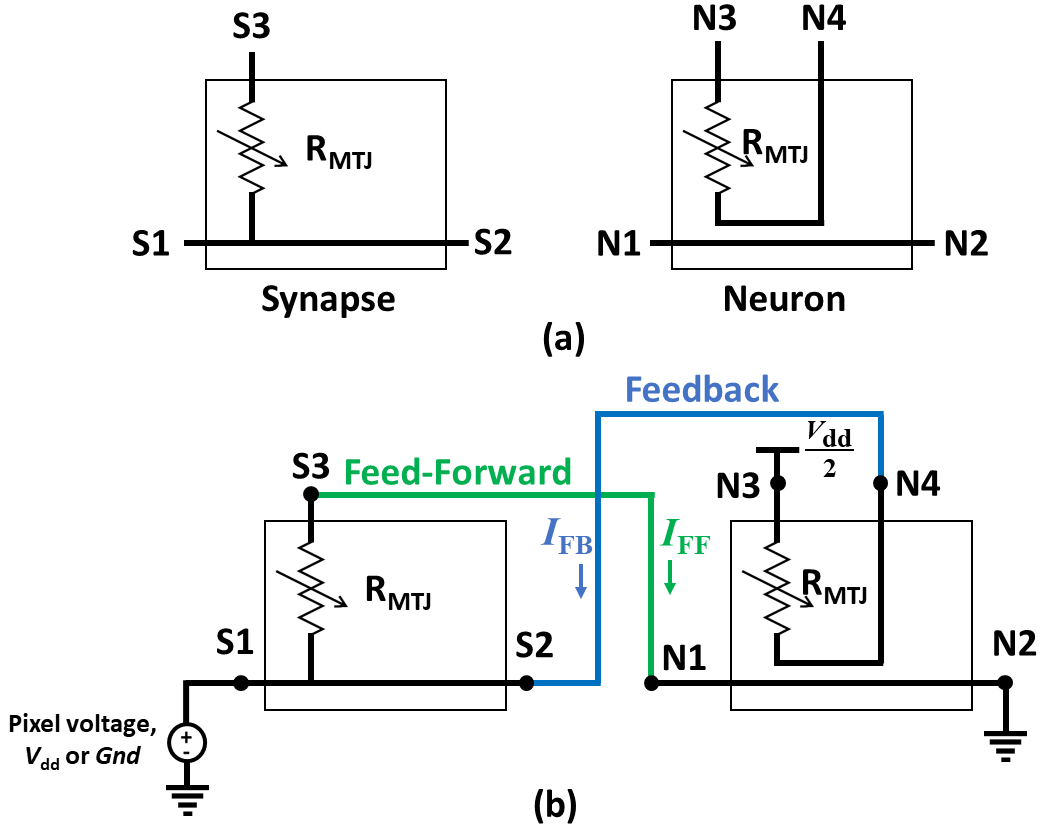} 
    \caption{(a) Electrical equivalent of the DW-MTJ synapse and neuron. (b) Synapse-Neuron connecting circuit. Feed-forward connection performs the inference operation and feedback connection performs the unsupervised learning.}
    \label{fig:Learning_Circuit}
\end{figure}

\section{Circuit for Unsupervised Learning}
The process of updating weights in a neural network is called learning. Depending on the presence of a teacher signal, learning can be classified into the categories of supervised, unsupervised and semi-supervised methods. In our hardware system, we have focused on optimizing the unsupervised learning primitive. This primitive can later be co-integrated with other learning sub-systems, as demonstrated in Section \ref{sec:learning_appl}. 
Our approach implements unsupervised learning as an electrical approximation of Hebbian (associative) learning, which can be efficiently implemented via the spike-timing-dependent plasticity (STDP) rule; this rule can efficiently extract information from real-time data, such as images, video, or other raw sensor data \cite{masquelier2007unsupervised}.

Fig. \ref{fig:Learning_Circuit}(a) shows the electrical equivalents of the DW-MTJ synapse and neuron discussed in section \ref{sec:Synapse_Neuron}. The synapse is represented by an analog resistor $R_\text{MTJ}$ in the S1-S3 path and a conducting wire in the S1-S2 path. The conducting wire represents the DW track which has a negligible resistance compared to the MTJ resistance. The neuron equivalent also consists of a conducting wire in the N1-N2 path representing the DW track and a binary resistor $R_\text{MTJ}$ in the N3-N4 path. The isolation between the N1-N2 and N3-N4 paths results from the electrically insulated magnetic coupling layer in Fig. \ref{fig:Synapse_Neuron} (\textit{b}).  

Fig. \ref{fig:Learning_Circuit}(b) illustrates the neuron-synapse connectivity underlying the inference and learning mechanisms. The synapse is connected between an input pattern source and a post-synaptic neuron. The image pixel can be converted into binary 0 and 1 input patterns corresponding to voltage levels $Gnd$ and $V_\text{dd}$, respectively. The input voltage is connected to the S1 terminal of the synapse and a feed-forward connection is made between the S3 and N1 terminals. The feed-forward current $I_\text{FF}$ flows through the S1-S3-N1-N2 path, is modulated by the analog resistor $R_\text{MTJ}$ corresponding to the synapse, and results in the integration process of the post-synaptic neuron. 

The learning is performed by creating a feedback connection between the N4 and S2 terminals. The feedback current $I_\text{FB}$ flows through the N3-N4-S2-S1 path and is modulated by the binary resistor $R_\text{MTJ}$ in the N3-N4 path. The N3 terminal is connected to voltage level $V_\text{dd}$/2. Therefore, the feedback current $I_\text{FB}$ has a positive (negative) value if the input voltage connected to the N1 terminal is $Gnd$ ($V_\text{dd}$). The positive (negative)  $I_\text{FB}$ current through the S2-S1 path decreases (increases) the synaptic analog conductance of $R_\text{MTJ}$ connected in the S1-S3 path, as mentioned in Section \ref{sec:Synapse_Neuron}. The conductance update summary is provided in Table \ref{tab:Learning_Circuit}, where the synapse conductance changes only when the post-synaptic neuron fires. The binary resistor $R_\text{MTJ}$ connected between N3-N4 controls the rate of the learning mechanism. For a firing (non-firing) neuron, $R_\text{MTJ}$ connected between N3-N4 provides a high (low) conductance and results in a higher (lower) value of the $I_\text{FB}$ current and learning rate.  

\begin{table}[]
\caption{Learning Circuit Summary}
\centering
\begin{tabular}{|c|c|c|}
\hline
\multicolumn{1}{|l|}{Input / Pixel Voltage} & $I_\text{FB} \text{(S2-S1)}$     & \multicolumn{1}{l|}{Synapse Conductance} \\ \hline
Gnd                                       & Positive & Decreases                                \\ \hline
$\textit{V}_\text{dd}$                                       & Negative & Increases                                \\ \hline
\end{tabular}
\label{tab:Learning_Circuit}
\end{table}

\section{DW-MTJ Perceptron with Unsupervised Learning}
\begin{figure}
    \centering
    \includegraphics[width=8cm]{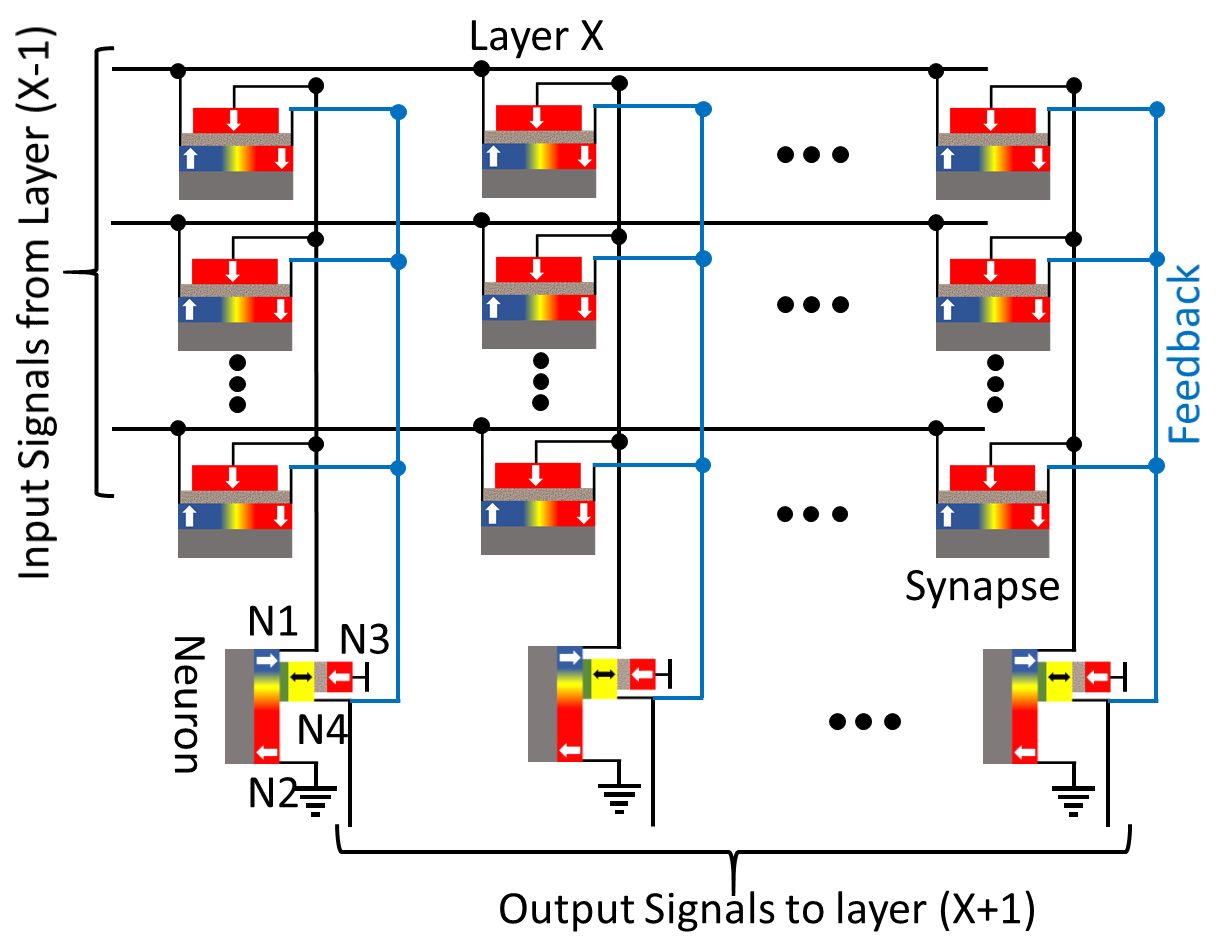} 
    \caption{3T DW-MTJ synapses arranged in a crossbar architecture. The output signals are a weighted product of the input signals times the conductance of their respective synaptic DW-MTJs. Weight update in this architecture is performed by the blue feedback lines from the 4T DW-MTJ neurons to the 3T DW-MTJ synapses.}
    \label{fig:MTJ}
    \label{fig:crossbar}
\end{figure}

Fig. \ref{fig:crossbar} shows the proposed DW-MTJ crossbar with a feedback connection from 4-terminal DW-MTJ neurons to 3-terminal DW-MTJ synapses in order to perform unsupervised learning. Input signals are provided to the left terminals of the synapses. Electrical signal flows across the tunnel barrier to the central terminal of each synapse, eventually reaching the connected output neuron. The current flowing across each synapse tunnel barrier is a function of the DW position, and therefore the synaptic weight given by the conductance of the DW-MTJ \cite{senguptaTbiocas}. The weighted current fed into the N1 terminal of the output neuron performs LIF functionalities by modulating its DW position. The feedback operation to perform unsupervised learning is achieved by connecting the N4-terminal of the output neuron to the right terminals of the synapses connected to its N1 terminal.

\section{Characteristic Equations}
\label{section:behaviors}
The behavior of DW-MTJ synapses and neurons is given by equations that characterize tunnel barrier conductance as a function of DW position and the motion of DWs as a function of current flow between the left and right terminals. For the synapse $s$, its conductance $G_s$ at time $t$ is given by (\ref{eqn:RS}), where $x^t_s$ is the position of the synapse DW along the track, $w_s$ is the width of the tunnel barrier, $G_\text{AP}$ is the anti-parallel conductance, and $G_\text{P}$ is the parallel conductance. For consistency with the literature on neural networks, we use this conductance value to define the weight of the synapse $\omega^t_s$ as a normalized value between $0$ and $1$ in (\ref{eqn:weight}). In the neuron $n$, the short tunnel barrier produces only binary conductance states characterized by (\ref{eqn:RN}), where $b_n$ is the position of the neuron barrier along the track. In every synapse $s$, a current $I$ causes the DW to move from its current position $x^t_s$ to $x^{t+1}_s$ after a time $\Delta T$ (\ref{eqn:XS}). In the neuron $n$, a leaking force $L$ is included such that the DW dynamics are given by (\ref{eqn:XN}). Due to this leaking and the lateral inhibition among nearby neurons, the perceptron system can be designed to ensure that exactly one neuron fires in response to each set of inputs, and that the neurons are reset before the next set of inputs is provided \cite{hassanJAP}.

\begin{gather}
    G^t_s = \frac{x^t_s}{w_s} G_\text{AP} + \frac{w_s - x^t_s}{w_s} G_\text{P} \in [G_\text{AP}, G_\text{P}]
\label{eqn:RS} \\ 
\omega^t_s = \frac{G^t_s-G_\text{AP}}{G_\text{P} - G_\text{AP}} \in [0, 1]
\label{eqn:weight} \\ 
\begin{aligned}
G^t_n = \begin{cases} G_\text{P} \hspace*{36pt} & x^t_n < b_n  \\ G_\text{AP} & x^t_n > b_n \end{cases}
\end{aligned}
\label{eqn:RN} \\ 
x^{t+1}_s = x^t_s + \Delta T \times \alpha(x^t_s, I)
\label{eqn:XS} \\ 
x^{t+1}_n = x^t_n + \Delta T  \times \left[ \alpha(x^t_n, I) - L(x^t_n) \right]
\label{eqn:XN}
\end{gather}






\section{Clustering Algorithm}
\label{sec:algo}

\begin{figure}
    \centering
    \includegraphics[width=\columnwidth]{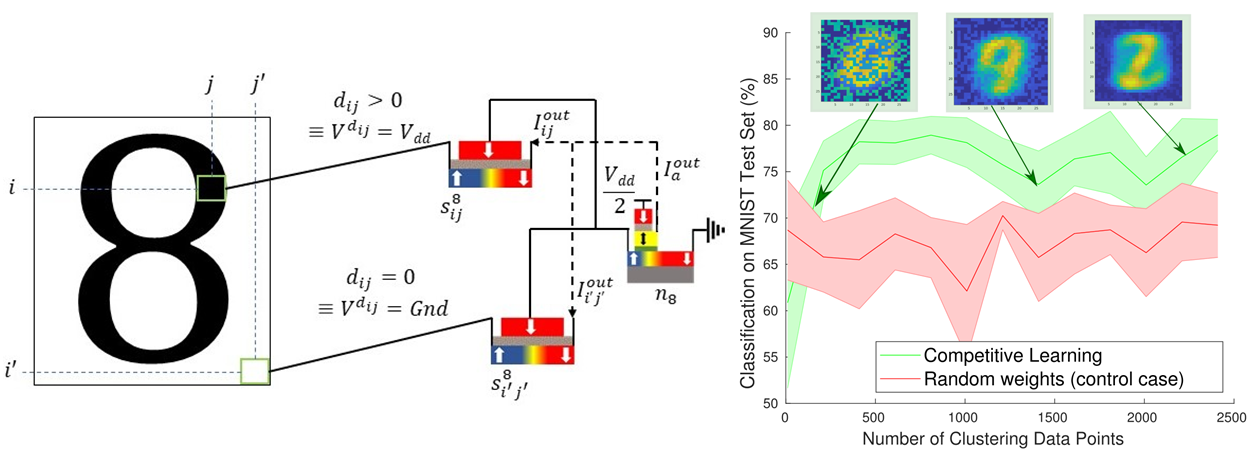} 
    \caption{(\textit{left}) During training, the output of the best-matching neuron is fed to the right terminal of every synapse connected to it. This changes the weights of said synapses in such a way that synapses which do not contribute to a positive signal are weakened and those that do are strengthened. (\textit{Right}) Test-set classification on the MNIST dataset is given as a function of datapoints presented to the system learning with clustered weights (green); as visible, this approach converges quickly and outpaces the constant weights system (red) not benefiting from this operation. The insets construct example receptive fields from the clustered system, showing the first phase where it is still converging and noisy and two later more stable cases.}
    \label{fig:learning}
    \label{fig:example2}
\end{figure}

%

\noindent \textbf{Given:} $m$ $n \times n$ images, $x$ fully connected independent layers with $n^2$ synapses $s^k_{ij}$ each connected to one output neuron $n_k$.

\noindent \textbf{Output:} Trained synaptic weights such that output neurons act as classifiers by assigning the input data to one of $x$ clusters.

\noindent \textbf{Procedure:} For every data point $d$, do the following, where $d_{ij} \in \{0, 1\}$ denotes pixel ($i, j$) as in Fig. \ref{fig:example2}:
\begin{enumerate}
    \item Inference: Feed $d$ to each layer by applying voltage $V_\text{dd}$ (Gnd) for pixels with value 1 (0). Let $I^\text{out}_a = \max_{k \in \{1, \dots, x\}} I^\text{out}_k$ be the largest neuron output current.
    \item Training: Feed $d$ to layer $a$  by applying $V_\text{dd}$ (Gnd) for pixels with value 1 (0). Use output current $I^\text{out}_{ij}$ as feedback to the right terminal of synapses $s^a_{ij}$. 
    \item $x^{t+1}_{s^a_{ij}} = x^t_{s^a_{ij}} + \Delta T \times \alpha \left( x^t_{s^a_{ij}}, V^{d_{ij}} \in \{\text{Gnd}, V_\text{dd}\}, I^\text{out}_{ij} \right)$
    \item Update conductance and weights $G^{t+1}_s, \omega^{t+1}_s$ 
    \item $t = t + 1$ (Procedure starts at $t=0$).
\end{enumerate}

Our approach relies on two phases: training and inference. In line (3): If $V^{d_{ij}} = \text{Gnd}$, then the feedback current $I^\text{out}_{ij}$ flows through synapse $s^a_{ij}$ in the right-to-left direction, leading to $\omega^{t+1}_{s^a_{ij}} < \omega^t_{s^a_{ij}}$. That is, the synaptic weight of $s^a_{ij}$ will be decreased because this synapse did not contribute to detecting pixel $d_{ij}$. If $V^{d_{ij}} = V_\text{dd}$, then the feedback current $I^\text{out}_{ij}$ flows in the left-to-right direction through the $s^a_{ij}$ synapse, leading to $\omega^{t+1}_{s^a_{ij}} \geq \omega^t_{s^a_{ij}}$. That is, the synaptic weight of $s^a_{ij}$ will be increased because this synapse contributed to detecting pixel $d_{ij}$. Since the neurons are reset after each input, we need not worry about the changes in $x^{t+1}_n, G^{t+1}_n$ for output neurons. 



\section{Learning and Recognition Results}
\label{sec:learning_appl}
\subsubsection{Integrated approach for on-chip classification}
Next, we combined the DW-MTJ clustering network (the encoder) with an appropriate read-out crossbar (the decoder) and tested it on the MNIST digit recognition challenge. This system reads spikes from the encoder and maps them to the labels provided at the output of the system, adjusting conductances following a supervised rule, as in \cite{bennettsemisupervised}. In this work, we apply the same overall method but adapt DW-MTJ  synapses in the first layer intrinsically, following Section \ref{section:behaviors}. We initially simulated the hard winner-take-all case, where lateral inhibition $L$ is tuned in (5) such that one neuron in the encoder fires at each training moment. The results of this demonstration are shown in Fig. \ref{fig:learning}, where performance on the dataset is plotted as a function of the total size of the set of datapoints $D$ presented for clustering, and compared to a control case where the encoder's weights are untrained. In both cases, $S=3e4$ supervised (labeled) examples are given to the decoder network to map spikes to labels, and $N_\text{HL}=250$ competing neurons cluster together. The clustering system alone receives $U=3e3$ samples prior to read-out and these are used to adapt the weights following the algorithm introduced in Section \ref{sec:algo}. As visible, while the algorithm at first performs worse than a random distribution of conductances, it quickly outperforms and reaches around $80\%$ classification on the task at $D=1000$ and thereafter. 

\begin{figure}
    \centering
    \includegraphics[width=\columnwidth]{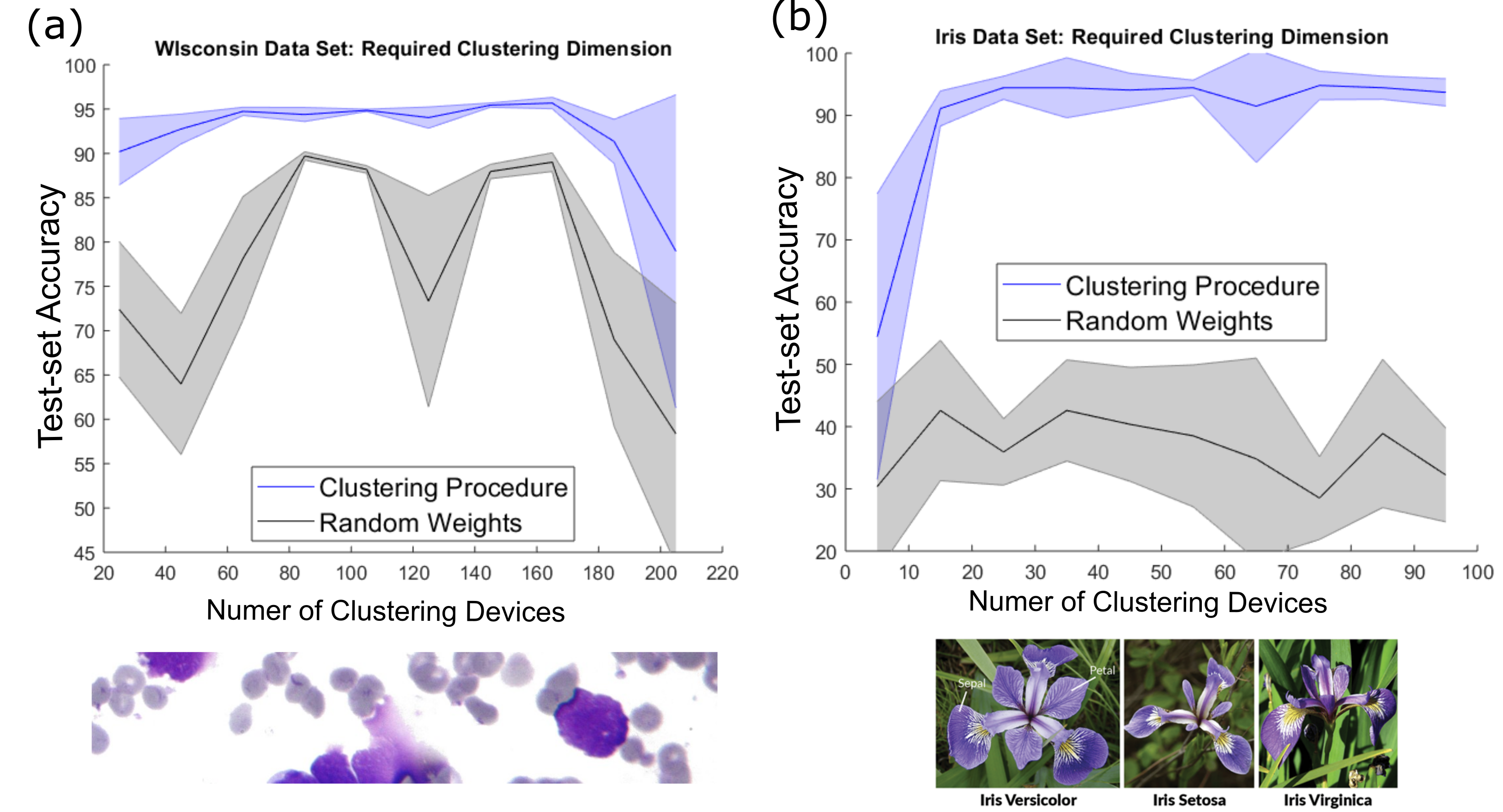} 
    \caption{(\textit{a}) Shows the performance in correctly identifying malignant or benign cells from the test-set of the Wisconsin task as a function of competing units in the unsupervised/random layer; (\textit{b}) shows the same results for the Iris test where a correct guess in the test-set consists of the correct species of Iris. For both tasks, $S=4e3$ supervised and $U=2e3$ unsupervised examples from the training set where shown (the latter were only presented to the systems exploited clustering).}
    \label{fig:cluster_new}
\end{figure}

\subsubsection{State-of-the-art performance on clustering tasks}
Since our algorithm is shallow, it may not be suited for challenging/large classification tasks such as the MNIST task. However, it is an exciting way to perform a natural clustering operation that can otherwise be performed in a small network such as a perceptron, support vector machine, or k-means clustering system. In order to demonstrate the utility of our algorithm, we attempt two tasks which have previously been used in clustering with memristive devices:
\begin{itemize}
    \item The Fisher IRIS task. This task consists of 150 examples of flower properties collected for three species of Iris plants, with each data point using 4 different type of measured flower characteristics.  In \cite{jeong2018k}, this task was solved with emerging devices using a centroid learning based approach, which achieved $93.3 \%$. However, this centroid approach requires complex search and update schemes. 
    \item The Wisconsin breast cancer dataset, which consists of 398 datapoints, each consisting of many cell properties (fractal dimension, concavity, area etc) observed during an assay. The objective is to determine if cells are cancerous or benign (\textit{e.g.}, the clustering/classification task is binary). This task was solved using an approximation for Principal Component Analysis (PCA) using Sanger's rule in a small crossbar with $97.6 \%$ accuracy. However, Sanger's rule requires a very complex pulse width calculation to implement, making it difficult to implement on-chip with high energy efficiency .
\end{itemize}

In Fig. \ref{fig:cluster_new}(a), we demonstrate an average  result of $96.94\%$ and best result of $98.11\%$ accuracy on the Wisconsin task when between $N_\text{HL}=40$ and $N_\text{HL}=160$ devices participate in clustering and they use the soft Winner-Take-All (WTA) formulation (\textit{e.g.}, multiple domain-wall neurons can fire at a given example/moment). These results assume a split of 227 training and 171 test points between the two cases, and compare  favorably to results obtained using a software clustering algorithm \cite{dubey2016analysis}.  Meanwhile, in the Fisher Iris task, average best result of $94.34\%$ and best result of $96.84\%$ is achieved for any amount of clustering units greater than $N_\text{HL}=20$ and again using the soft-WTA approach (Fig. \ref{fig:cluster_new}(a)).

In both cases, we have also contrasted to a case where the clustering layer is forced to random weights, so as to demonstrate that our system has more computational power than a perceptron alone. Note that the pictured random weights performance is somewhat restricted from best possible performances ($91.14\%$ for Iris, $93.9\%$ for Wisconsin) as we have restricted the total number of supervised learning samples to $S=4e3$ in both cases. This highlights the fast learning capabilities provided via intrinsic clustering operations and physically enabled by the DW-MTJ device properties. Additionally, since the clustering performance we have achieved is better than hyperplane classification boundaries have when fully converged (\textit{e.g.}, the standard perceptron read-out), we confirm that our clustering layer captures a meaningful latent statistical interpretation of the tasks. As suggested in \cite{tavanaei2018training}, our clustered system may contain Hidden Markov Model (HMM) representations and implement an approximate version of the Expectation Maximization (EM) algorithm.
\section{Conclusion}
In this work, we highlight how a set of DW-MTJ synapses connected to several competitively learning DW-MTJ neurons can implement an effective form of unsupervised learning (Clustering) and have demonstrated the fundamentals of the electrical circuit and algorithm implementation that make this possible. Through concrete use-cases on the MNIST, Fisher Iris, and Wisconsin Breast Cancer datasets, we have demonstrated the utility of this approach and its superiority to standard shallow neural network learning. In our next work, we plan to build upon these promising early results by incorporating additional bio-plausible neuronal effects such as homeostasis and resonant and fire, on the neuron level, and three-factor and other additional plasticity effects at the synapse level. 




\bibliographystyle{IEEEtran}
\bibliography{main}

\end{document}